%% file: neurips_data_2023.tex
\documentclass{article}

\usepackage[numbers]{natbib}
    \usepackage[preprint]{neurips_data_2023}

\usepackage[utf8]{inputenc} %
\usepackage[T1]{fontenc}    %
\usepackage[hidelinks]{hyperref}       %
\usepackage{url}            %
\usepackage{booktabs}       %
\usepackage{amsfonts}       %
\usepackage{nicefrac}       %
\usepackage{microtype}      %
\usepackage{graphicx}
\usepackage{amsmath}
\usepackage{tikz}
\usepackage{color}
\usetikzlibrary{shapes}
\usepackage{multirow}
\usepackage{longtable}
\usepackage{wrapfig,lipsum,booktabs}
\usepackage{adjustbox}
\usepackage[frozencache=true,cachedir=minted-cache]{minted} 
\definecolor{LightGray}{gray}{0.9}

\newcommand{\mailto}[1]{\href{mailto:#1}{\texttt{#1}}}
\newcommand{\name}{\texttt{ClimateLearn}}
\title{ClimateLearn: Benchmarking Machine Learning for Weather and Climate Modeling}

\author{
    Tung Nguyen\thanks{Equal contribution.} \\
    UCLA \\
    \mailto{tungnd@cs.ucla.edu}
    \And
    Jason Jewik\footnotemark[1] \\
    UCLA \\
    \mailto{jason.jewik@ucla.edu}
    \And
    Hritik Bansal \\
    UCLA \\
    \mailto{hbansal@ucla.edu}
    \And
    Prakhar Sharma \\
    UCLA \\
    \mailto{prakhar6sharma@gmail.com} \\
    \And
    Aditya Grover \\
    UCLA \\
    \mailto{adityag@cs.ucla.edu}
}

\begin{document}

\maketitle

\input{sections/abstract}           %
\input{sections/intro}              %
\input{sections/related_work}       %
\input{sections/key_features}       %
\input{sections/benchmarks}
%
\input{sections/conclusion}
\input{sections/acknowledgements}

\newpage 
\bibliography{references}
\bibliographystyle{plainnat}

\include{sections/appendix}

\end{document}

%% file: sections/abstract.tex
\begin{abstract}
Modeling weather and climate is an essential endeavor to understand the near- and long-term impacts of climate change, as well as inform technology and policymaking for adaptation and mitigation efforts. 
In recent years, there has been a surging interest in applying data-driven methods based on machine learning for solving core problems such as weather forecasting and climate downscaling.
Despite promising results, much of this progress has been impaired due to the lack of large-scale, open-source efforts for reproducibility, resulting in the use of inconsistent or underspecified datasets, training setups, and evaluations by both domain scientists and artificial intelligence researchers. 
We introduce \name{}, an open-source PyTorch library that vastly simplifies the training and evaluation of machine learning models for data-driven climate science.
\name{} consists of holistic pipelines for dataset processing (e.g., ERA5, CMIP6, PRISM), implementation of state-of-the-art deep learning models (e.g., Transformers, ResNets), and quantitative and qualitative evaluation for standard weather and climate modeling tasks.
We supplement these functionalities with extensive documentation, contribution guides, and quickstart tutorials to expand access and promote community growth.
We have also performed comprehensive forecasting and downscaling experiments to showcase the capabilities and key features of our library.
To our knowledge, \name{} is the first large-scale, open-source effort for bridging research in weather and climate modeling with modern machine learning systems.
Our library is available publicly at \url{https://github.com/aditya-grover/climate-learn}.
\end{abstract}

%% file: sections/intro.tex
\section{Introduction}
\label{intro}

The escalating extent, duration, and severity of extreme weather events such as droughts, floods, and heatwaves in recent decades are some of the most devastating outcomes of climate change.
Moreover, as average surface temperature is anticipated to continue rising through the end of the century, such extreme weather events are likely to occur with even greater intensity and frequency in the future~\cite{Coughlan2023,IPCC2021,Martinich2019,Moon2021,Rodell2023,ClimateGov2020}.
Two key devices used by scientists to understand historical trends and make such predictions about future weather and climate are the general circulation model (GCM) and the numerical weather prediction (NWP) model.
These models represent Earth system components including the atmosphere, land surface, ocean, and sea ice as intricate dynamical systems, and they are the prevailing paradigm for weather and climate modeling today due to their established reliability, well-founded design, and extensive study~\cite{Bauer2015,Danabasoglu2020,Hersbach2020,Neelin2010}.
However, they also suffer from notable limitations such as inadequate resolution of several subgrid processes, coarse representation of local geographical features, incapacity to utilize sources of observational data (e.g., weather stations, radar, satellites), and substantial demand for computing resources~\cite{Balaji2017,Lavers2022,Leung2003,Rauscher2010}. 
These deficiencies combined with the expanding availability of petabyte-scale climate data~\cite{Copernicus,Eyring2016,NASAEarthdata} and lowering compute requirements of machine learning (ML) models in recent years have motivated researchers from both the climate science and artificial intelligence (AI) communities to investigate the application of ML-based methods in weather and climate modeling \cite{Bochenek2022,deBurghDay2023,Karpatine2019,Kashinath2021,McGovern2017,Mosavi2018,Prudden2020,Sundararajan2021,Willard2020}.

In spite of this growing interest, the improvements have been marred by the lack of practically grounded data benchmarks, open-source model implementations, and transparency in evaluation.
For example, many papers in weather forecasting~\cite{Bi2022,Brust2021,Espeholt2022,Grover2015,Keisler2022,Pathak2022,Ravuri2021,Sonderby2020,Tekin2021,Weyn2019,Weyn2020,Zhang2019}, climate projection~\citep{Watson-Parris2022}, and climate downscaling~\cite{Bano2020,Liu2020,Nagasato2021,Rodrigues2018,Sachindra2018,Vandal2019} choose to benchmark on different geographical regions, temporal ranges, evaluation metrics, and data augmentation strategies.
These inconsistencies can confound the source of reported improvements and promotes a culture of irreproducible scientific practices~\cite{Kapoor2022}. 
Recently, there have been some leaderboard benchmarks, such as WeatherBench~\cite{Rasp2020}, ClimateBench~\cite{Watson-Parris2022}, and FloodNet~\cite{rahnemoonfar2021floodnet},
 that propose datasets and baselines for specific tasks in climate science, but a holistic software ecosystem that encompasses the entire data, modeling, and evaluation pipeline across several tasks is lacking. 

To bridge this gap, we propose \name{}, an open-source, user-friendly PyTorch library for data-driven climate science.
To the best of our knowledge, it is the first software package to provide end-to-end ML pipelines for weather and climate modeling. 
\name{} supports data pre-processing utilities, implements popular deep learning models along with traditional baseline methods, and enables easy quantification and visualization of data and model predictions for fundamental tasks in climate science, including weather forecasting, downscaling, and climate projections.
One segment of \name{}'s target user demographic are weather and climate scientists, who possess expertise in the physical laws and phenomena relevant for constructing robust modeling priors, but might lack familiarity with optimal approaches for implementing, training, and evaluating machine learning models. 
Another segment of \name{}'s target user demographic are ML researchers, who might encounter difficulties framing weather and climate modeling problems in a scientifically sound and practically useful manner, working with climate datasets---which is quite heterogeneous and often exists in bespoke file formats uncommon in mainstream ML research (e.g., NetCDF), or appropriately quantifying and visualizing their results for interpretation and deployments.

To showcase the capabilities of \name{} and establish benchmarks, we perform and report results from numerous experiments on the supported tasks with a variety of traditional methods and our own tuned implementations of deep learning models on weather and climate datasets. 
In addition to traditional evaluation setups, we have created novel dataset and benchmarking scenarios to test model robustness and applicability to forecasting extreme weather events.
 Further, the library is modular and easily extendable to include additional tasks,  datasets, models, metrics, and visualizations.
 We have also provided extensive documentation and contribution guides for improving the ease of community adoption and accelerating open-source expansion.
 While our library is already public, we are releasing all our data, code, and model checkpoints for the benchmarked evaluation in this paper to aid reproducibility and broader interdisciplinary research efforts.

%% file: sections/related_work.tex
\section{Related work}
\label{related_work}

Recent works have proposed benchmark datasets for weather and climate modeling problems. Prominently, \citet{Rasp2020} proposed WeatherBench, a dataset for weather forecasting based on ERA5, followed by an extension called WeatherBench Probability~\cite{Garg2022}, which adds support for probabilistic forecasting.
\citet{mouatadid2021learned} extend similar benchmarks to the subseasonal to seasonal timescale.
For precipatation-events such as rain specifically, there are prior datasets such as RainBench~\citep{de2021rainbench} and IowaRain~\cite{sit2021iowarain}.
There exist datasets such as ExtremeWeather~\citep{Racah2017}, FloodNet~\citep{rahnemoonfar2021floodnet},  EarthNet~\citep{requena2021earthnet2021}, DroughtED~\citep{minixhofer2021droughted} and ClimateNet~\citep{Prabhat2021} for detection and localization of extreme weather events, and NADBenchmarks~\citep{proma2022nadbenchmarks} for natural disasters related tasks. \citet{cachay2021climart} recently proposed ClimART, a benchmark dataset for emulating atmospheric radiative transfer in weather and climate models.
For identifying long-term, globally-averaged trends in climate, \citet{Watson-Parris2022} proposed ClimateBench, a dataset for climate model emulation.

Beyond plain datasets, libraries such as Scikit-downscale~\cite{Hammon2022}, CCdownscaling~\cite{Polasky2022}, and CMIP6-Downscaling~\cite{CarbonPlan2022} provide tools for post-processing of climate model outputs via statistical, non-deep-learning downscaling, or mapping low-resolution gridded, image-like inputs to high-resolution gridded outputs.
In a slightly different approach, pyESD focuses on downscaling from gridded climate data to specific weather stations~\cite{Boateng2023}.
 Pyrocast~\citep{tazi2022pyrocast} proposes an integrated ML pipeline to forecast Pyrocumulonimbus (PyroCb) Clouds.
Many of these works supply only individual components of an ML pipeline but do not always have an API for loading climate data into a ML-ready format, or standard model implementations and evaluation protocols across multiple climate science tasks.
As an end-to-end ML pipeline, \name{} holistically bridges the gap for applying ML to challenging weather and climate modeling tasks like forecasting, downscaling, and climate projection.

%% file: sections/key_features.tex
\section{Key Components of \name{}}
\label{key_features}
\input{figures/features}

\name{} is a PyTorch library that implements a range of functionalities for benchmarking of ML models for weather and climate. Broadly, our library is comprised of four components: tasks, datasets, models, and  evaluations.
See Figure~\ref{fig:1} for an illustration.
Sample code snippets for configuring each component is provided in Appendix~\ref{sec:code_snippets}.

\subsection{Tasks}
\label{tasks}

\textbf{Weather forecasting } is the task of predicting the weather at a future time step $t + \Delta t$ given the weather conditions at the current step $t$ and optionally steps preceding $t$. A ML model receives an input of shape $C \times H \times W$ and predicts an output of shape $C' \times H \times W$.
$C$ and $C'$ denote the number of input and output channels, respectively, which contain variables such as geopotential, temperature, and humidity.
$H$ and $W$ together denote the spatial coverage and resolution of the data, which depend on the region studied and how densely we grid it.
In our benchmarking, we focus on forecasting at the global scale, but \name{} can be easily extended to regional forecasting.

\textbf{Downscaling } Due to their high computational cost, existing climate models often use large grid cells, leading to low-resolution predictions.
While useful for understanding large-scale climate trends, these do not provide sufficient detail to analyze local phenomena and design regional policies.
The process of correcting biases in climate model outputs and mapping them to higher resolutions is known as downscaling.
ML models for downscaling are trained to map an input of shape $C \times H \times W$ to a higher resolution output $C' \times H' \times W'$, where $H' > H$ and $W' > W$. As in forecasting, in downscaling, $H \times W$ and $H' \times W'$ can span either the entire globe or a specific region.

\textbf{Climate projection } aims to obtain long-term predictions of the climate under different forcings, e.g., greenhouse gas emissions. We provide support to download data from ClimateBench~\cite{Watson-Parris2022}, a recent benchmark designed for testing ML models for climate projections. Here, the task is to predict the annual mean distributions of $4$ climate variables: surface temperature, diurnal temperature range, precipitation, and the $90$th percentile of precipitation, given four anthropogenic forcing factors: carbon dioxide (CO$_2$), sulfur dioxide (SO$_2$), black carbon (BC), and methane (CH$_4$). 

\subsection{Datasets}
\label{data}

\textbf{ERA5 } is a commonly-used data source for training and benchmarking data-driven forecasting and downscaling methods~\citep{Bi2022,Lam2022,Nagasato2021,Nguyen2023,Pathak2022,Rasp2020,Rasp2021}. It is maintained by the European Center for Medium-Range Weather Forecasting (ECMWF)~\cite{Hersbach2020}. ERA5 is a reanalysis dataset that provides the best guess of the state of the atmosphere and land-surface variables at any point in time by combining multiple sources of observational data with the forecasts of the current state-of-the-art forecasting model known as the Integrated Forecasting System (IFS)~\cite{Wedi2015}. In its raw format, ERA5 contains hourly data from $1979$ to the current time on a $0.25^\circ$ grid of the Earth's sphere, with different climate variables at $37$ different pressure levels plus the Earth’s surface. This corresponds to nearly $400{,}000$ data points with a resolution of $721 \times 1440$. As this data is too big for most deep learning models, \name{} also supports downloading a smaller version of ERA5 from WeatherBench~\cite{Rasp2020}, which uses a subset of ERA5 climate variables and regrids the raw data to lower resolutions.

\textbf{Extreme-ERA5} is a subset of ERA5 that we have constructed to evaluate forecasting performance in extreme weather situations. Specifically, we consider ``simple extreme'' events \cite{Watson2001}, i.e., weather events that have individual climate variables exceeding critical values locally. Heat waves and cold spells are examples of extreme events that can be quantitatively captured by extreme localized surface-level temperatures over prolonged days. To mimic real-world scenarios, we calculate thresholds for each pixel of the grid using the $5$th and $95$th percentile of the $7$-day localized mean surface temperature over the training period ($1979$-$2015$). We then select a subset of pixels from all the available pixels in the testing set (2017-18) that had a $7$-day localized mean surface temperature beyond these thresholds. We refer to Appendix~\ref{sec:era5-extreme} for more details.

\textbf{CMIP6 } is a collection of simulated data from the Coupled Model Intercomparison Project Phase 6 (CMIP6)~\cite{Eyring2016}, an international effort across different climate modeling groups to compare and evaluate their global climate models. While the main goal of CMIP6 is to improve the understanding of Earth's climate systems, the data from their experimental runs is freely accessible online. CMIP6 data covers a wide range of climate variables, including temperature and precipitation, from hundreds of climate models, providing a rich source of data. For our forecasting experiments, we specifically use the ouputs of CMIP6's MPI-ESM1.2-HR model, as it contains similar climate variables to those represented in ERA5 and was also considered in previous works for pretraining deep learning models~\cite{Rasp2021}. MPI-ESM1.2-HR provides data from $1850$ to $2015$ with a temporal resolution of 6 hours and spatial resolution of $1^\circ$. Since this again corresponds to a grid that is too big for most deep learning models, we provide lower resolution versions of this dataset for training and evaluation.
Besides, we also perform experiments with ClimateBench, which contains data on a range of future emissions scenarios based on simulations by the Norwegian Earth System Model~\cite{Seland2020}, another member of CMIP6.
We refer to Appendix~\ref{sec:cmip6} for time ranges and more details of the experiments.

\textbf{PRISM } is a dataset of various observed atmospheric variables like precipitation and temperature over the conterminous United States at varying spatial and temporal resolutions from 1895 to present day. 
It is maintained by the PRISM Climate Group at Oregon State University~\cite{PRISM}.
At the highest publicly available resolution, PRISM contains daily data on a grid of 4 km by 4 km cells (approximately $0.03^\circ$), which corresponds to a matrix of shape $621\times 1405$.
For the same reason we regrid ERA5 and CMIP6, we also provide a regridded version of raw PRISM data to $0.75^\circ$ resolution.

\subsection{Models}
\label{models}

\textbf{Traditional baselines } 
\name{} provides the following traditional baseline methods for forecasting: climatology, persistence, and linear regression.
The climatology method uses historical average values of the predictands as the forecast.
In \name{}, we consider the climatology of a particular variable to be its average value over the entire training set.
The persistence method uses the last observed values of the predictands as the forecast.
For downscaling, \name{} provides nearest and bilinear interpolation. Nearest interpolation estimates the value of an unknown pixel to be the value of the nearest known pixel. Bilinear interpolation estimates the value at an unknown pixel by taking the weighted average of neighboring pixels.

\textbf{Deep learning models}
The data for gridded weather and climate variables is represented as a 3D matrix, where latitude, longitude, and the variables form the height, width, and channels, respectively. Hence, convolutional neural networks (CNNs) are commonly used for forecasting and downscaling, which can be viewed as instances of the image-to-image translation problem~\cite{Espeholt2022,Hess2022,Nagasato2021,Rasp2021,Rodrigues2018,Singh2021,Sonderby2020,Tekin2021,Vandal2017,Weyn2019,Weyn2020}. \name{} supports ResNet~\cite{He2015} and U-Net~\cite{Ronneberger2015}---two prominent variants of the commonly used CNN architectures. Additionally, \name{} supports Vision Transformer (ViT)~\cite{Bi2022,Gao2023,Nguyen2023}, a class of models that represent images as a sequence of pixel patches. \name{} also supports loading benchmark models from the literature such as \citet{Rasp2021} in a single line of code and is built so that custom models can be added easily.

\subsection{Evaluations}
\label{metrics}

\textbf{Forecasting metrics } 
For deterministic forecasting, \name{} provides metrics such as root mean square error (RMSE) and anomaly correlation coefficient (ACC), which measures how well model forecasts match ground truth anomalies. 
For probabilistic forecasting, \name{} provides spread-skill ratio and continuous ranked probability score, as defined by \citet{Garg2022}. 
\name{} also provides latitude-weighted version of these metrics, which lends extra weight to pixels near the equator.
This is needed because the curvature of the Earth means that grid cells at low latitudes cover less area than grid cells at high latitudes.
We refer to Appendix \ref{sec:metrics} for additional details, including equations.

\textbf{Downscaling metrics }
For downscaling, \name{} uses RMSE, mean bias, and Pearson's correlation coefficient, in which mean bias is the difference between the spatial mean of ground-truth values and the spatial mean of predictions.
We refer to Appendix \ref{sec:metrics} for additional details.

\textbf{Climate projection metrics } In addition to the standard RMSE metric, we provide two metrics suggested by ClimateBench: Normalized spatial root mean square error (NRMSE$_s$) and Normalized global root mean square error (NRMSE$_g$). 
We refer to Appendix~\ref{sec:metrics} for more details.

\textbf{Visualization }
Besides these quantitative evaluation procedures, \name{} also provides ways for users to inspect model performance qualitatively through visualizations of data and model predictions. 
For instance, in a single line of code, users can visually inspect their forecasting model's per-pixel mean bias, or the expected values of forecast errors, over the testing period. 
Such a visualization can be useful for pinpointing the regions on which the model's predictions consistently deviate from the ground truth in a certain direction. 
For probabilistic forecasts, \name{} can generate the corresponding rank histogram, which indicates the reliability and sharpness of the model. 
Sample visualizations of deterministic and probabilistic predictions are provided in Appendix \ref{sec:visualizations}.

%% file: figures/features.tex
\begin{figure}[t]
    \centering
    \begin{adjustbox}{width=0.8\textwidth}
    \begin{tikzpicture}
        \draw[black, fill=white, rounded corners=10pt] (0,0) rectangle (0.99\textwidth,9.8);
        \draw[dotted] (\textwidth/2,0) -- (\textwidth/2,9.8);
        \draw[dotted] (0,5) -- (\textwidth,5);

        \node[rounded corners=5pt, draw, fill=blue!20, text centered, minimum width=3cm, minimum height=0.75cm] at (3.4, 9.25) {Datasets \& Benchmarks};
        \node[rounded corners, text=black, inner sep=5pt] (node2) at (2.9, 6.4) {\textbf{Various Physical Variables}};
  \node[inner sep=0pt, outer sep=0pt, anchor=east, right = 0.002cm] at (node2.east) {\color{black!30!green}\checkmark};
          \node[rounded corners, text=black, inner sep=5pt] (node1) at (2.75, 5.9) {\textbf{Fast Dataloading}};
  \node[inner sep=0pt, outer sep=0pt, anchor=east, right = 0.002cm] at (node1.east) {\color{black!30!green}\checkmark};
  \node[rounded corners, text=black, inner sep=5pt] (node2) at (2.8, 5.4) {\textbf{Preprocessed Datasets}};
  \node[inner sep=0pt, outer sep=0pt, anchor=east, right = 0.002cm] at (node2.east) {\color{black!30!green}\checkmark};

    \node[rounded corners=5pt, draw, fill=blue!20, text centered, minimum width=3cm, minimum height=0.75cm] at (0.75\textwidth, 9.25) {Tasks};
 \node[rounded corners=5pt, draw, fill=red!10, text centered, minimum width=1cm, minimum height=0.5cm] at (10.5, 8.25) {Forecasting, Projections, Downscaling};

        \node[rounded corners, text=black, inner sep=5pt] (node2) at (10.5, 6.4) {\textbf{Various Grid Resolutions}};
  \node[inner sep=0pt, outer sep=0pt, anchor=east, right = 0.002cm] at (node2.east) {\color{black!30!green}\checkmark};
   \node[rounded corners, text=black, inner sep=5pt] (node2) at (10.5, 5.9) {\textbf{
   Flexible Spatiotemporal Setups}};
  \node[inner sep=0pt, outer sep=0pt, anchor=east, right = 0.002cm] at (node2.east) {\color{black!30!green}\checkmark};
   \node[rounded corners, text=black, inner sep=5pt] (node2) at (10.5, 5.4) {\textbf{Extreme Events}};
  \node[inner sep=0pt, outer sep=0pt, anchor=east, right = 0.002cm] at (node2.east) {\color{black!30!green}\checkmark};

\node[rounded corners=5pt, draw, fill=blue!20, text centered, minimum width=3cm, minimum height=0.75cm] at (3.4, 4.4) {Models};
\node[rounded corners=5pt, draw, fill=red!10, text centered, minimum width=1cm, minimum height=0.5cm] at (1.7, 3.4) {Baselines};
\node[rounded corners=5pt, draw, fill=red!10, text centered, minimum width=1cm, minimum height=0.5cm] at (4.5, 3.4) {Deep Learning};
        \node[rounded corners, text=black, inner sep=5pt] (node2) at (3.2, 1.3) {\textbf{End-to-End Training Pipeline}};
  \node[inner sep=0pt, outer sep=0pt, anchor=east, right = 0.002cm] at (node2.east) {\color{black!30!green}\checkmark};
          \node[rounded corners, text=black, inner sep=5pt] (node1) at (3.3, 0.8) {\textbf{PyTorch Model Implementations}};
  \node[inner sep=0pt, outer sep=0pt, anchor=east, right = 0.002cm] at (node1.east) {\color{black!30!green}\checkmark};
  \node[rounded corners, text=black, inner sep=5pt] (node2) at (3.4, 0.3) {\textbf{Easy Customization and Tuning}};
  \node[inner sep=0pt, outer sep=0pt, anchor=east, right = 0.002cm] at (node2.east) {\color{black!30!green}\checkmark};

\node[rounded corners=5pt, draw, fill=blue!20, text centered, minimum width=3cm, minimum height=0.75cm] at (0.75\textwidth, 4.4) {Evaluation};
 \node[rounded corners=5pt, draw, fill=red!10, text centered, minimum width=1cm, minimum height=0.5cm] at (9, 3.4) {Metric Logging};
        \node[rounded corners=5pt, draw, fill=red!10, text centered, minimum width=1cm, minimum height=0.5cm] at (12, 3.4) {Visualizations};

           \node[rounded corners, text=black, inner sep=5pt] (node2) at (10.5, 1.3) {\textbf{Point-wise and Summary Statistics}};
  \node[inner sep=0pt, outer sep=0pt, anchor=east, right = 0.002cm] at (node2.east) {\color{black!30!green}\checkmark};
          \node[rounded corners, text=black, inner sep=5pt] (node1) at (10.4, 0.8) {\textbf{Error and Correlation Metrics}};
  \node[inner sep=0pt, outer sep=0pt, anchor=east, right = 0.002cm] at (node1.east) {\color{black!30!green}\checkmark};
  \node[rounded corners, text=black, inner sep=5pt] (node2) at (10.3, 0.3) {\textbf{Uncertainity Quantification}};
  \node[inner sep=0pt, outer sep=0pt, anchor=east, right = 0.002cm] at (node2.east) {\color{black!30!green}\checkmark};

    \node[inner sep=0pt] (image1) at (2,8.25) {\includegraphics[scale=0.4]{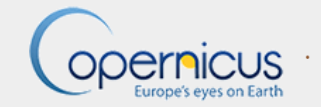}};
    \node[inner sep=0pt] (image2) at (5,8.25) {\includegraphics[scale=0.3]{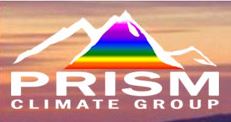}};
    \node[inner sep=0pt] (image3) at (3.5,7.25) {\includegraphics[scale=0.3]{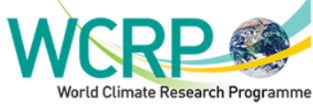}};

    \node[inner sep=0pt] (image5) at (9,7.25) {\includegraphics[scale=0.25]{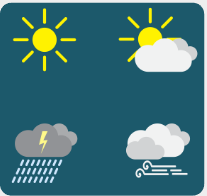}};
    \node[inner sep=0pt] (image6) at (12,7.25) {\includegraphics[scale=0.25]{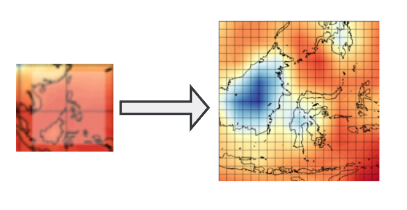}};

\node[inner sep=0pt] (image7) at (1.8,2.25) {\includegraphics[scale=0.15]{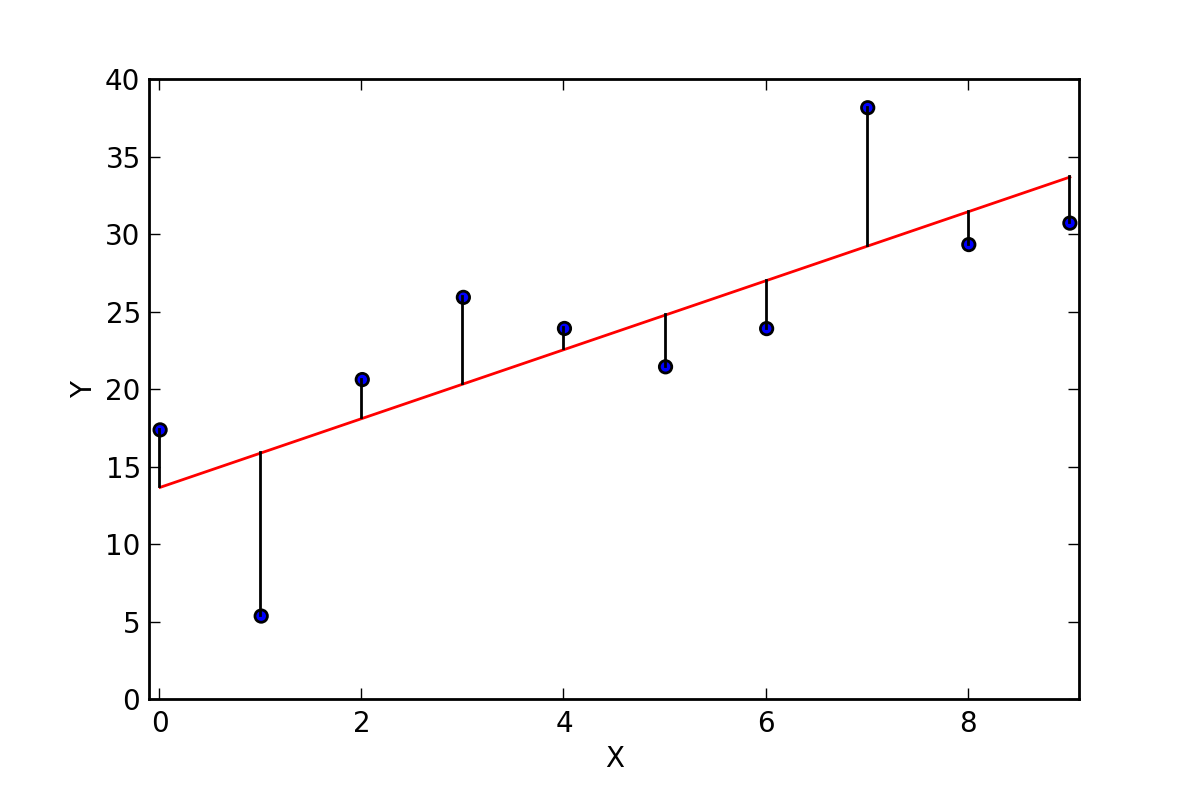}};
    \node[inner sep=0pt] (image8) at (4.6,2.25) {\includegraphics[scale=0.18]{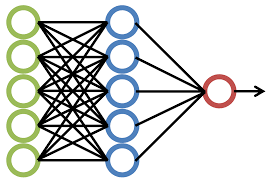}};

\node[inner sep=0pt] (image9) at (8.8,2.2) {\includegraphics[scale=0.06]{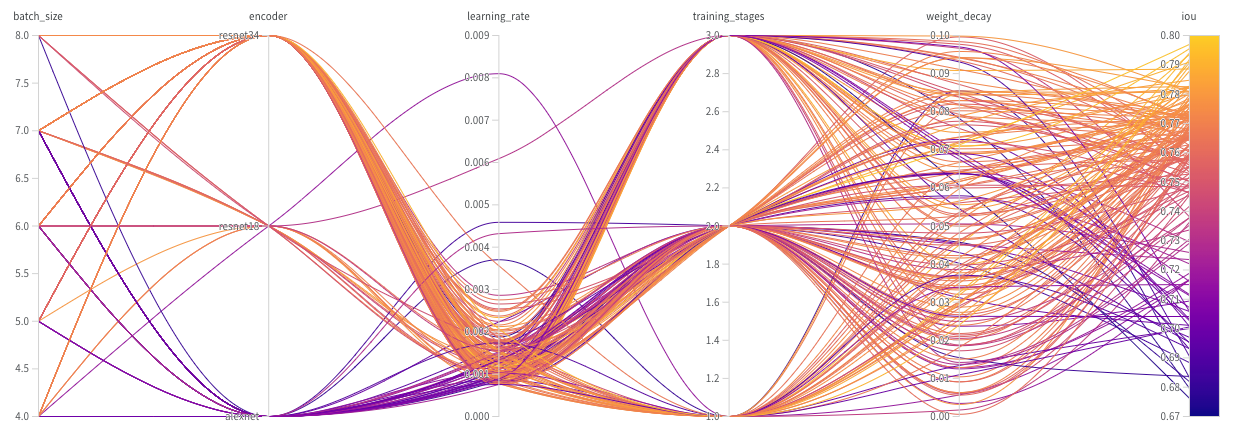}};
    \node[inner sep=0pt] (image10) at (12,2.2) {\includegraphics[scale=0.12]{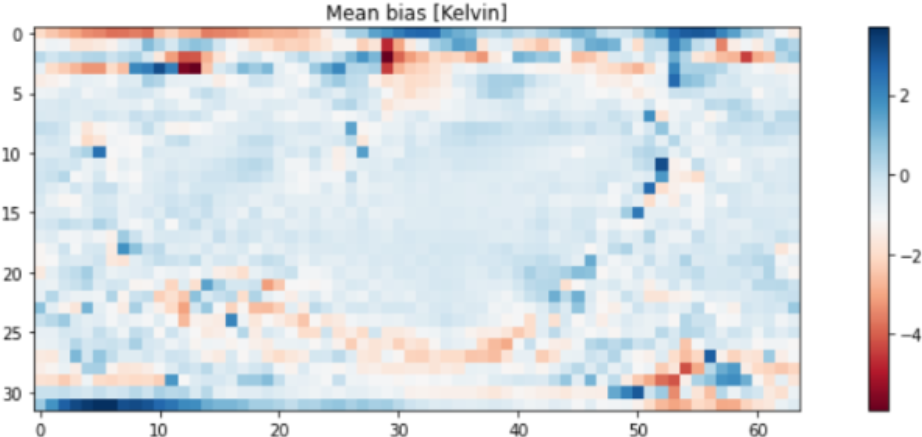}};

    \node[anchor=center] at (\textwidth/2,5) {\includegraphics[scale = 0.15]{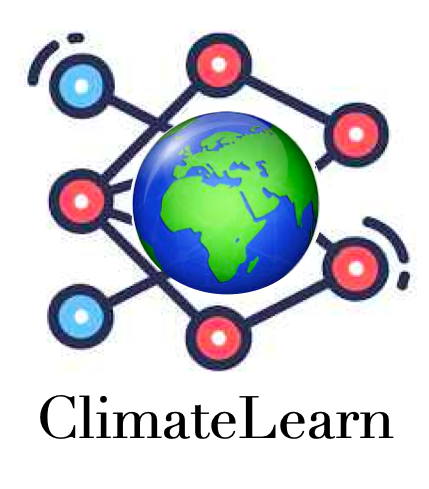}}; 
    
    \end{tikzpicture}
    \end{adjustbox}
    \caption{Key components of \name{}. We support observational, simulated, and reanalysis datasets from a variety of sources. The currently supported tasks are weather forecasting, downscaling, and climate projection. \name{} also provides a suite of standard baselines and deep learning architectures, along with common metrics, visualizations, and logging support.}
    \label{fig:1}
\end{figure}

%% file: sections/benchmarks.tex
\section{Benchmark Evaluation via \name{}}
\label{benchmarks}
In this section, we evaluate the performance of different deep learning methods supported by \name{} on weather forecasting and climate downscaling. We refer to Appendix~\ref{sec:climate_proj} for experiments on the climate projection task. We conduct extensive experiments and analyses with different settings to showcase the features and flexibility of our library.

\subsection{Weather forecasting} \label{sec:forecast_benchmark}
\begin{figure}[t]
    \centering
    \includegraphics[width=0.85\textwidth]{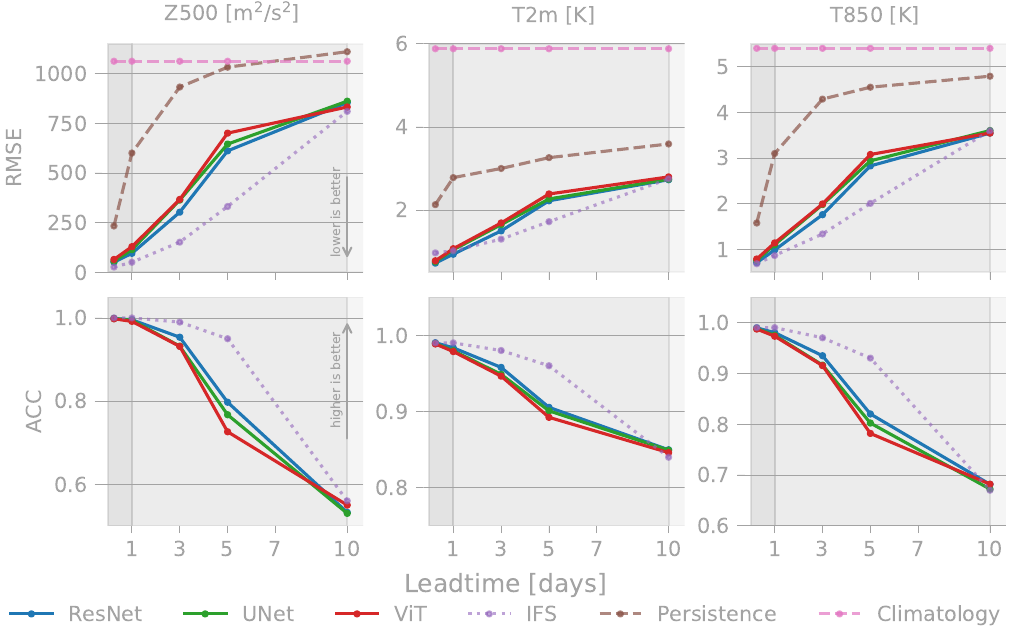}
    \caption{Performance on forecasting three variables at different lead times. Solid lines are deep learning methods, dashed lines are simple baselines, and the dotted line is the physics-based model. Lower RMSE and higher ACC indicate better performance.}
    \label{fig:forecast_benchmark}
\end{figure}

We first benchmark on weather forecasting. In addition, we compare different approaches for training forecast models in Section~\ref{sec:forecast_approaches}, and investigate the robustness of these models to extreme weather events and data distribution shift in Section~\ref{sec:extreme_events} and~\ref{sec:data_robustness}, respectively.

\textbf{Task } We consider the task of forecasting the geopotential at $500$hPa (Z500), temperature at $850$hPa (T850), and temperature at $2$ meters from the ground (T2m) at five different lead times: $6$ hours, and $\{1, 3, 5, 10\}$ days. Z500 and T850 are often used for benchmarking in previous works~\citep{Bi2022,Lam2022,Nguyen2023,Pathak2022,Rasp2021,Rasp2020}, while the surface variable T2m is relevant to human activities.

\textbf{Baselines } We consider ResNet~\cite{He2015}, U-Net~\cite{Ronneberger2015}, and ViT~\cite{Dosovitskiy2021} which are three common deep learning architectures in computer vision. We provide the architectural details of these networks in Appendix~\ref{sec:network_arc}. We perform direct forecasting, where we train one neural network for each lead time. In addition, we compare the deep learning methods with climatology, persistence, and IFS~\citep{Wedi2015}.

\textbf{Data } We use ERA5~\cite{Hersbach2020} at $5.625^{\circ}$ for training and evaluation, which is equivalent to having a $32 \times 64$ grid for each climate variable. 
The input variables to the deep learning models include geopotential, temperature, zonal and meridional wind, relative humidity, and specific humidity at $7$ pressure levels $(50, 250, 500, 600, 700, 850, 925)$hPa, $2$-meter temperature, $10$-meter zonal and meridional wind, incoming solar radiation, and finally $3$ constant fields: the land-sea mask, orography, and the latitude, which together constitute $49$ input variables. For non-constant variables, we use data at $3$ timesteps $t$, $t-6$h, and $t-12$h to predict the weather at $t + \Delta t$, resulting in $46 \times 3 + 3 = 141$ input channels. Each channel is standardized to have $0$ mean and $1$ standard deviation. The training period is from $1979$ to $2015$, validation in $2016$, and test in $2017$ and $2018$.

\textbf{Training and evaluation } We use latitude-weighted mean squared error as the loss function. We use AdamW optimizer~\cite{Loshchilov2019} with a learning rate of $5\times10^{-4}$ and weight decay of $1\times10^{-5}$, a linear warmup schedule for $5$ epochs, followed by cosine-annealing for $45$ epochs. We train for $50$ epochs with $128$ batch size, and use early stopping with a patience of $5$ epochs. We use latitude-weighted root mean squared error (RMSE) and anomaly correlation coefficient (ACC) as the test metrics.

\begin{figure}[t]
    \centering
    \includegraphics[width=0.85\textwidth]{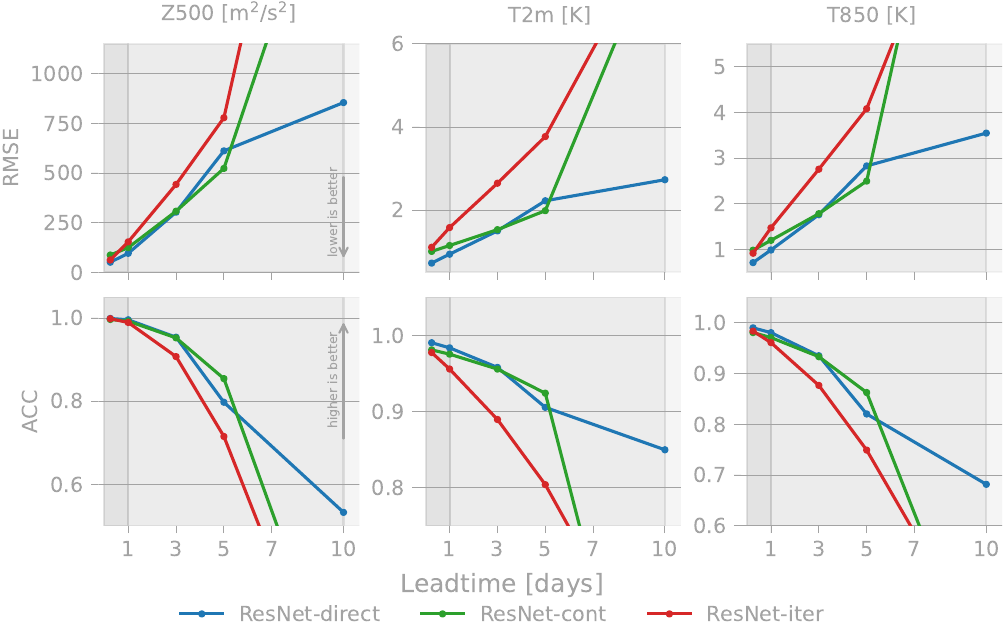}
    \caption{Comparison of direct, continuous, and iterative forecasting with ResNet architecture.}
    \label{fig:forecast_ablation}
\end{figure}

\textbf{Benchmark results} 
Figure~\ref{fig:forecast_benchmark} shows the performance of different baselines. As expected, the forecast quality in terms of both RMSE and ACC of all baselines worsens with increasing lead times. The deep learning methods significantly outperform climatology and persistence but underperform IFS. ResNet is the best-performing deep learning model on most tasks in both metrics. We hypothesize that while being more powerful than ResNet in general, U-Net tends to perform better when trained on high-resolution data~\citep{Ronneberger2015}, and ViT often suffers from overfitting when trained from scratch~\cite{He2022,Nguyen2023}. Our reported performance of ResNet closely matches that of previous works~\cite{Rasp2021}. 

\subsubsection{Should we perform direct, continuous, or iterative forecasting?} \label{sec:forecast_approaches}

In direct forecasting, we train a separate model for each lead time. This can be computationally expensive as the training cost scales linearly with the number of lead times. In this section, we consider two alternative approaches, namely, continuous forecasting and iterative forecasting, and investigate the trade-off between computation and performance.
In continuous forecasting, a model conditions on lead time information to make corresponding predictions, which allows the same trained model to make forecasts at any lead times. We refer to Appendix~\ref{sec:continuous_setup} for details on how to do this. In iterative forecasting, we train the model to forecast at a short lead time, i.e., $6$ hours, and roll out the predictions during evaluation to make forecasts at longer horizons. We note that in order to roll out more than one step, the model must predict all variables in the input. This provides the benefit of training a single model that can predict at any lead time that is a multiplication of $6$.

We compare the performance of direct, continuous, and iterative forecasting using the same ResNet architecture with training and evaluation settings identical to Section~\ref{sec:forecast_benchmark}. Figure~\ref{fig:forecast_ablation} shows that ResNet-cont slightly underperforms the direct model at $6$-hour and $1$-day lead times, but performs similarly or even better at $3$-day and $5$-day forecasting. We hypothesize that for difficult tasks, training with randomized lead times enlarges the training data and thus improves the generalization of the model. A similar result was observed by~\citet{Rasp2020}. However, the continuous model does not generalize well to unseen lead times, which explains the poor performance when evaluated at $10$-day forecasting. ResNet-iter performs the worst in the three approaches, which achieves a reasonable performance at $6$-hour lead time, but the prediction error accumulates exponentially at longer horizons. This was also observed in previous works~\citep{Nguyen2023,Rasp2020}. We believe this issue can be mitigated by multi-step training~\citep{Pathak2022}, which we leave as future work.

\subsubsection{Extreme weather prediction} \label{sec:extreme_events}

\begin{wraptable}{r}{0.45\textwidth}
\vspace{-0.25in}
    \centering
    \caption{Latitude-weighted RMSE on the normal and extreme test splits of ERA5.}
    \label{tab:extreme_era5_forecasting}
    \resizebox{0.4\textwidth}{!}{\begin{tabular}{@{}lccc@{}}
    \toprule
    T2M & 6 Hours     & 1 Day       & 3 Days\\ \midrule
    Climatology     & $5.87$ / $6.51$ & $5.87$ / $6.53$ & $5.87$ / $6.58$ \\
    Persistence     & $2.76$ / $2.99$ & $2.13$ / $1.78$ & $2.99$ / $2.42$ \\
    ResNet          & $0.72$ / $\textbf{0.72}$ & $0.94$ / $\textbf{0.91}$ & $1.50$ / $\textbf{1.33}$ \\
    U-Net            & $0.76$ / $0.77$ & $1.04$ / $0.99$ & $1.65$ / $1.43$ \\
    ViT            & $0.78$ / $0.80$  & $1.09$ / $1.05$ & $1.71$ / $1.55$ \\ \bottomrule
    \end{tabular}}
\end{wraptable}

Despite the surface-level temperature being an outlier, Table \ref{tab:extreme_era5_forecasting} shows that deep learning and persistence perform better on Extreme-ERA5 than ERA5 for all different lead times. Climatology performs worse on Extreme-ERA5, which is expected since predicting the mean of the target distribution is not a good strategy for outlier data. The persistence performance indicates that the variation between the input and output values is comparatively less for extreme weather conditions. We hypothesize that, although the marginal distribution $p(y)$ of the subset data is extreme, the conditional distribution $p(y|x)$ which we are trying to model is not extreme. Thus, we are not experiencing any drop in performance on such a subset. While from a ML standpoint, it might seem necessary to evaluate the models for cases where the conditional distribution is extreme for any input variable, such a dataset might not qualify under the well-known categories of extreme events. Future studies  for constructing extreme datasets could try targeting extreme events such as floods, which are usually caused by high amounts of precipitation under a short period of time~\cite{Stephenson2008}.

\subsubsection{Data robustness of deep learning models} \label{sec:data_robustness}

\begin{wraptable}{r}{0.4\textwidth}
\vspace{-0.25in}
    \caption{Performance of ResNet trained on one dataset (columns) and evaluated on another (rows).}
    \label{tab:robustness_forecasting}
    \resizebox{0.4\textwidth}{!}{
\begin{tabular}{@{}clcccc@{}}
\toprule
                   &      & \multicolumn{2}{c}{ERA5}                           & \multicolumn{2}{c}{CMIP6}                          \\ \midrule
\multicolumn{1}{l}{}   &      & \multicolumn{1}{l}{ACC} & \multicolumn{1}{l}{RMSE} & \multicolumn{1}{l}{ACC} & \multicolumn{1}{l}{RMSE} \\ \midrule
\multirow{3}{*}{ERA5}  & Z500 & $\textbf{0.95}$           & $\textbf{322.86}$          & $0.93$                    & $345.00$                   \\
                   & T850 & $\textbf{0.93}$           & $\textbf{1.90}$            & $0.90$                    & $2.21$                     \\
                   & T2m  & $\textbf{0.95}$           & $\textbf{1.62}$            & $0.93$                    & $1.94$                     \\ \midrule
\multirow{3}{*}{CMIP6} & Z500 & $0.95$                    & $357.66$                    & $\textbf{0.96}$           & $\textbf{306.86}$           \\
                   & T850 & $0.91$                    & $2.11$                     & $\textbf{0.94}$           & $\textbf{1.70}$            \\
                   & T2m  & $0.93$                    & $1.91$                     & $\textbf{0.96}$           & $\textbf{1.53}$            \\ \bottomrule
\end{tabular}}
\vspace{-0.2in}
\end{wraptable}

We study the impact of data distribution shifts on forecasting performance. We consider CMIP6 and ERA5 as two different data sources. The input variables are similar to the standard setting, except that we remove relative humidity, $10$-meter zonal and meridional wind, incoming solar radiation, and the $3$ constant fields due to their unavailability in CMIP6. To account for differences in the temporal resolution and data coverage, we set the temporal resolution to $6$ hours and set $1979$-$2010$, $2011$-$12$, and $2013$-$14$ as training, validation, and testing years respectively.

Table \ref{tab:robustness_forecasting} shows that all methods achieve better evaluation scores if the training and testing splits come from the same dataset, but cross-dataset performance is not far behind, highlighting the robustness of the models across distributional shifts. We see a similar trend for different models across different lead times, which we refer to Appendix~\ref{sec:dataset_robustness} for more details. We also conducted an experiment where the years $1850$-$1978$ are included in training for CMIP6. The results show that for all models across almost all lead times, training on CMIP6 leads to even better performance on ERA5 than training on ERA5. 
For exact numbers and setup refer to Appendix~\ref{sec:dataset_robustness}.

\begin{table}[b]
\centering
\caption{Downscaling experiments on ERA5 ($5.6^{\circ}$) to ERA5 ($2.8^{\circ}$) and ERA5 ($2.8^{\circ}$) to PRISM ($0.75^{\circ}$). For ERA5 ($5.6^{\circ}$) to ERA5 ($2.8^{\circ}$), Pearson's correlation coefficient was 1.0 for all models.}
\label{tab:downscaling_exp}
\resizebox{0.9\textwidth}{!}{%
\begin{tabular}{@{}lcccccc|ccc@{}}
\toprule
         & \multicolumn{6}{c|}{ERA5 to ERA5}                                                                          & \multicolumn{3}{c}{ERA5 to PRISM}              \\ \midrule
         & \multicolumn{2}{c}{Z500 (m$^2$ s$^{-2}$)} & \multicolumn{2}{c}{T850 (K)}  & \multicolumn{2}{c|}{T2m (K)}   & \multicolumn{3}{c}{Daily Max T2m (K)}          \\ \midrule
         & RMSE                 & Mean bias          & RMSE          & Mean bias     & RMSE          & Mean bias     & RMSE          & Mean bias      & Pearson       \\ \midrule
Nearest  & $269.67$               & $\textbf{0.04}$      & $1.99$          & $\textbf{0.00}$ & $3.11$          & $\textbf{0.00}$ & $2.91$          & $\textbf{-0.05}$ & $0.89$          \\
Bilinear & $134.07$               & $\textbf{0.04}$      & $1.50$          & $\textbf{0.00}$ & $2.46$          & $\textbf{0.00}$ & $2.64$          & $0.12$           & $0.91$          \\
ResNet   & $54.20$                & $-6.41$              & $\textbf{0.39}$ & $-0.05$         & $\textbf{1.10}$ & $-0.22$         & $1.86$          & $-0.11$          & $0.95$          \\
Unet     & $\textbf{43.84}$       & $-6.55$              & $0.94$          & $-0.06$         & $\textbf{1.10}$ & $-0.12$         & $\textbf{1.57}$ & $-0.14$          & $\textbf{0.97}$ \\
ViT      & $85.32$                & $-35.98$             & $1.03$          & $-0.01$         & $1.25$          & $-0.20$         & $2.18$          & $-0.26$          & $0.94$          \\ \bottomrule
\end{tabular}%
}
\end{table}

\subsection{Downscaling}

\textbf{Task and data }
We consider two settings for downscaling. In the first setting, we downscale 5.625$^\circ$ ERA5 data to 2.8125$^\circ$ ERA5, both at a global scale and hourly intervals.
The input and target variables are the same as used in Section \ref{sec:forecast_benchmark}.
In the second setting, we consider downscaling 2.8125$^\circ$ ERA5 data over the conterminous United States to 0.75$^\circ$ PRISM data over the same region at daily intervals. This is equivalent to downscaling a reanalysis/simulated dataset to an observational dataset, similar to previous papers~\cite{Hanssen2005,Rodrigues2018,Tang2018}.
The cropped ERA5 data has shape $9\times 21$ while the regridded PRISM data is padded with zeros to the shape $32 \times 64$.
The only input and output variable is daily max T2m, which is normalized to have $0$ mean and $1$ standard deviation.
The training period is from 1981 to 2015, validation is in 2016, and the testing period is from 2017 to 2018.

\textbf{Baselines }
We compare ResNet, U-Net, and ViT with two baselines: nearest and bilinear interpolation.

\textbf{Training and evaluation }
We use MSE as the loss function with the same optimizer and learning rate scheduler as in Section~\ref{sec:forecast_benchmark}, with an initial learning rate of $1\times 10^{-5}$.
A separate model is trained for each output variable,
and all models post-process the results of bilinear interpolation. 
We use RMSE, Pearson's correlation coefficient, and mean bias as the test metrics.
All metrics are masked properly since PRISM does not have data over the oceans.
See Appendix \ref{sec:metrics} for further details.

\textbf{Benchmark results }
Table \ref{tab:downscaling_exp} shows the performance of different baselines in both settings. As expected for the first setting, all methods achieve relatively low errors. The deep learning models outperformed both interpolation methods significantly on RMSE, but tend to overestimate the target variables, leading to negative mean bias. In the second setting---where the input and output come from two different datasets---the performance of all baselines drops. Nonetheless, the deep learning models again outperform the baseline methods on RMSE and exhibit negative mean bias, but also achieve higher Pearson's correlation coefficients. 

%% file: sections/conclusion.tex
\section{Conclusion}
\label{conclusion}
We presented \name{}, a user-friendly and open-source PyTorch library for data-driven weather and climate modeling.
Given the pressing nature of climate change, we believe our contribution is timely and of potential use to both the ML and climate science communities.
Our objective is to provide a standardized benchmarking platform for evaluating ML innovations in climate science, which currently suffer from challenges in standardization, accessibility, and reproducibility. \name{} provides users access to all essential components of end-to-end ML pipeline, including data pre-processing utilities, ML model implementations, and rigorous evaluations via metrics and visualizations. 
We use the flexible and modular design of \name{} to design and perform diverse experiments comparing deep learning methods with relevant baselines on our supported tasks. 

\textbf{Limitations and Future Work } 
In this work, we highlighted key features of the \name{} library, encompassing datasets, tasks, models, and evaluations. However,
we acknowledge that there are numerous avenues to enhance the comprehensiveness of our library in each of these dimensions.
One such avenue involves integrating regional datasets and expanding the catalog of available data sources.
On the modeling side, we plan to develop efficient implementations for training ensembles in service of critical uncertainty quantification efforts. 
In future iterations of our library, we will also integrate a hub of large-scale pretrained neural networks specifically designed for weather and climate applications~\cite{Bi2022,Lam2022,Nguyen2023,Pathak2022}.
Once integrated, these pretrained models will be further customizable through fine-tuning, enabling straightforward adaptation to downstream tasks.
Furthermore, we plan to incorporate support for physics-informed neural networks and other hybrid baselines that amalgamate physical models with machine learning methods, which will allow users to leverage the strengths of both paradigms.
Ultimately, our overarching objective is to establish \name{} as a trustworthy AI development tool for weather and climate applications.

%% file: sections/acknowledgements.tex
\begin{ack}
We thank Shashank Goel, Jingchen Tang, Seongbin Park, Siddharth Nandy, and Sri Keerthi Bolli for their contributions to \name{}. Aditya Grover was supported in part by a research gift from Google. Hritik Bansal was supported in part by AFOSR MURI grant FA9550-22-1-0380.
\end{ack}

%% file: sections/appendix.tex
\appendix

\section{Licenses and Terms of Use} \label{sec:license}

\name{} is a software package that can be installed from the Python Package Index as follows.
\begin{minted}{bash}
    pip install climate-learn
\end{minted}
The source code is available online under the MIT License at \url{https://github.com/aditya-grover/climate-learn}, and the accompanying documentation website is at \url{https://climatelearn.readthedocs.io/}. The Extreme-ERA5 dataset does not exist as a distinct entity, but can be produced by running code provided in our library. The Machine Intelligence Group at UCLA is the maintainer of \name{}.

The sources for datasets provided by \name{} are WeatherBench, ClimateBench, the Earth System Grid Federation (ESGF), the Copernicus Climate Data Store (CDS), and PRISM. The WeatherBench dataset (\url{https://mediatum.ub.tum.de/1524895}), ClimateBench dataset (\url{https://zenodo.org/record/7064308}), and MPI-ESM1.2-HR outputs from ESGF (\url{https://pcmdi.llnl.gov/CMIP6/TermsOfUse/TermsOfUse6-2.html}) are available under the CC BY 4.0 license. Neither Copernicus (\url{https://cds.climate.copernicus.eu/api/v2/terms/static/licence-to-use-copernicus-products.pdf}) nor PRISM (\url{https://prism.oregonstate.edu/terms/}) use a Creative Commons License. Instead, they each set forth their own terms of use, both of which permit reproduction and distribution for non-commercial purposes.

\section{Experiment details}
\label{sec:experiment_details}

\subsection{Network architectures} \label{sec:network_arc}
\subsubsection{ResNet}
Our ResNet architecture is similar to that of WeatherBench~\citep{Rasp2020,Rasp2021}, in which each residual block consists of two identical convolutional modules: 2D convolution $\rightarrow$ LeakyReLU with $\alpha = 0.3$ $\rightarrow$ Batch Normalization $\rightarrow$ Dropout. 

\begin{table}[h!]
\centering
\caption{Default hyperparameters of ResNet}
\label{tab:resnet_hyper}
\begin{tabular}{@{}lll@{}}
\toprule
Hyperparameter   & Meaning                                          & Value \\ \midrule
Padding size     & Padding size of each convolution layer           & $1$   \\
Kernel size      & Kernel size of each convolution layer            & $3$   \\
Stride           & Stride of each convolution layer                 & $1$   \\
Hidden dimension & Number of output channels of each residual block & $128$ \\
Residual blocks  & Number of residual blocks                        & $28$  \\
Dropout          & Dropout rate                                     & $0.1$ \\ \bottomrule
\end{tabular}
\end{table}

Table~\ref{tab:resnet_hyper} shows the hyperparameters for ResNet in all of our experiments. We use a convolutional layer with a kernel size of $7$ at the beginning of the network. All paddings are periodic in the longitude direction and zeros in the latitude direction.

\subsubsection{UNet}
We borrow our UNet implementation from \href{https://github.com/microsoft/pdearena/blob/main/pdearena/modules/twod_unet.py}{PDEArena}~\citep{gupta2022towards}. 
Table~\ref{tab:unet_hyper} shows the hyperparameters for UNet in all of our experiments. Similar to ResNet, we use a convolutional layer with a kernel size of $7$ at the beginning of the network, and all paddings are periodic in the longitude direction and zeros in the latitude direction.
\begin{table}[h!]
\centering
\caption{Default hyperparameters of UNet}
\label{tab:unet_hyper}
\begin{tabular}{@{}lll@{}}
\toprule
Hyperparameter          & Meaning                                                                                                  & Value          \\ \midrule
Padding size            & Padding size of each convolution layer                                                                   & $1$            \\
Kernel size             & Kernel size of each convolution layer                                                                    & $3$            \\
Stride                  & Stride of each convolution layer                                                                         & $1$            \\
Hidden dimension & Base number of output channels & $64$ \\
Channel multiplications & \begin{tabular}[c]{@{}l@{}}Determine the number of output channels\\ for Down and Up blocks\end{tabular} & $[1, 2, 2]$ \\
Blocks                  & Number of blocks                                                                                         & $2$            \\
Use attention           & If use attention in Down and Up blocks                                                                   & False          \\
Dropout                 & Dropout rate                                                                                             & $0.1$          \\ \bottomrule
\end{tabular}
\end{table}

\newpage
\subsubsection{ViT}
We use the standard Vision Transformer architecture~\citep{Dosovitskiy2021} with minor modifications. We remove the class token and add a 1-hidden MLP prediction head which is applied to the tokens after the last attention layer to predict the outputs. Tabel~\ref{tab:vit_hyper} shows the hyperparameters for ViT in all of our experiments.
\begin{table}[h]
\centering
\caption{Default hyperparameters of ViT}
\label{tab:vit_hyper}
\begin{tabular}{@{}lll@{}}
\toprule
Hyperparameter   & Meaning                                                                                                  & Value                                                                                      \\ \midrule
$p$              & Patch size & $2$ \\
$D$              & Embedding dimension                                                                                      & $128$                                                                                     \\
Depth            & Number of ViT blocks                                                                                     & $8$                                                                                        \\
\# heads         & Number of attention heads                                                                                & $4$                                                                                       \\
MLP ratio        & \begin{tabular}[c]{@{}l@{}}Determine the hidden dimension of\\ the MLP layer in a ViT block\end{tabular} & $4$                                                                                        \\
Prediction depth & Number of layers of the prediction head                                                                  & $2$                                                                                        \\
Hidden dimension & Hidden dimension of the prediction head                                                                  & $128$                                                                                     \\
Drop path        & For stochastic depth~\citep{huang2016deep}                                                               & $0.1$                                                                                      \\
Dropout          & Dropout rate                                                                                             & $0.1$                                                                                      \\ \bottomrule
\end{tabular}
\end{table}

\subsection{Datasets}
\subsubsection{ERA5} \label{sec:era5}
We refer to {\url{https://confluence.ecmwf.int/display/CKB/ERA5\%3A+data+documentation}} for more details of the raw ERA5 data.
We use the preprocessed version of ERA5 at $5.625^\circ$ from WeatherBench~\citep{Rasp2020} for our experiments.
Table~\ref{tab:era5_data} summarizes the variables we use for our experiments.
\begin{table}[h]
    \centering
    \caption{{ERA5 variables used in our experiments}. 
    \emph{Constant} represents constant variables, \emph{Single}  represents surface variables, and \emph{Atmospheric} represents atmospheric properties at the chosen altitudes.}    
    \label{tab:era5_data}
    \begin{tabular}{rllll}
        \toprule
        Type & Variable name & Abbrev. & Levels
        \\
        \midrule
         Static & Land-sea mask & LSM & \\
         Static & Orography & & & \\
         Static & Latitude & & & \\
         Single & Toa incident solar radiation & Tisr & \\
         Single & 2 metre temperature & T2m & \\
         Single & 10 metre U wind component & U10 & \\
         Single & 10 metre V wind component & V10 & \\
         \midrule
         Atmospheric & Geopotential & Z & 50, 250, 500, 600, 700, 850, 925 \\
         Atmospheric & U wind component & U & 50, 250, 500, 600, 700, 850, 925 \\
         Atmospheric & V wind component & V & 50, 250, 500, 600, 700, 850, 925 \\
         Atmospheric & Temperature & T & 50, 250, 500, 600, 700, 850, 925 \\
         Atmospheric & Specific humidity & Q & 50, 250, 500, 600, 700, 850, 925 \\
         Atmospheric & Relative humidity & R & 50, 250, 500, 600, 700, 850, 925 \\
        \bottomrule
    \end{tabular}
\end{table}

\subsubsection{Extreme-ERA5} \label{sec:era5-extreme}
\textbf{Calculating thresholds } We use the surface temperature (T2m) data corresponding to the years $1979-2015$ from ERA5 at a resolution of $5.625^\circ$ to calculate the thresholds. The thresholds are localized i.e. they are calculated for every pixel on the grid. For a given timestamp and pixel, we first calculate a 7 day mean till that timestamp. Now, to account for neighboring regions/pixels, we set the localized mean as $0.44$ * current pixel's mean $+$ $0.11$ * sum of means of pixels sharing an edge $+$ $0.027$ * sum of means of pixels sharing a vertex but not an edge. Note, that there is no need of padding while accounting for neighboring pixels, since earth is a globe, the neighbors of leftmost pixels include the rightmost pixels and vice-versa. Finally, the $5$th and $95$th percentile values of this new mean data corresponding to every pixel is set as threshold.

\textbf{Building masks } As the purpose of Extreme-ERA5 is evaluation of forecasting models under extreme weather conditions, we build it for test years i.e. $2017-2018$ only. We first create a $2$-D mask of size, latitude x longitude, filled with zeros for every available timestamp in the test years. Similar to the calculating thresholds, we compute the mean of each pixel at every timestamp for T2m's test data. We then, set the value for a given pixel in the mask as 1, if the mean value is outside the bounds set by the thresholds. Finally, during evaluation time, we use these masks to select subset of data.

\subsubsection{CMIP6} \label{sec:cmip6}
\textbf{MPI-ESM1.2-HR }
We use MPI-ESM1.2-HR, a dataset in the CMIP6 data repository for our experiments in Section~\ref{sec:data_robustness}. Table~\ref{tab:mpi_data} summarizes the variables we use for our experiments.
\begin{table}[h]
    \centering
    \caption{{MPI-ESM1.2-HR variables used in our experiments}. 
    \emph{Single}  represents surface variables and \emph{Atmospheric} represents atmospheric properties at the chosen altitudes.}    
    \label{tab:mpi_data}
    \begin{tabular}{rllll}
        \toprule
        Type & Variable name & Abbrev. & Levels
        \\
        \midrule
         Single & 2 metre temperature & T2m & \\
         Single & 10 metre U wind component & U10 & \\
         Single & 10 metre V wind component & V10 & \\
         \midrule
         Atmospheric & Geopotential & Z & 50, 250, 500, 600, 700, 850, 925 \\
         Atmospheric & U wind component & U & 50, 250, 500, 600, 700, 850, 925 \\
         Atmospheric & V wind component & V & 50, 250, 500, 600, 700, 850, 925 \\
         Atmospheric & Temperature & T & 50, 250, 500, 600, 700, 850, 925 \\
         Atmospheric & Specific humidity & Q & 50, 250, 500, 600, 700, 850, 925 \\
        \bottomrule
    \end{tabular}
\end{table}

\textbf{ClimateBench } 
We adopt data from ClimateBench~\citep{Watson-Parris2022} for our climate projection experiment. ClimateBench contains simulated data from experimental runs by the Norwegian Earth System Model~\citep{Seland2020}, a member of CMIP6, on different emission scenarios. Specifically, ClimateBench includes $7$ esmission scenarios: historical, ssp126, ssp370, ssp585, hist-aer, hist-GHG, and ssp245. We refer to the original ClimateBench paper for the exact temporal coverage and more details of these scenarios.

\subsection{Training details} \label{sec:training_details}

\subsubsection{Continuous training } \label{sec:continuous_setup}
Continuous models additionally condition on lead times to make predictions. To do this, we add the lead time value in hours divided by $100$ to the input channels to make the model aware of the lead time it is forecasting at. During training, we randomize the lead time from $6$ hours to $5$ days $\Delta t \sim \mathcal{U}[6, 120]$, and during evaluation, we fix the lead time to a certain value to evaluate the model's performance at a certain lead time. This setting was commonly used in previous works~\cite{Nguyen2023,Rasp2020}.

\subsubsection{Software and hardware stack}
We use PyTorch \citep{pytorch2019}, \texttt{numpy}~\citep{harris2020array} and \texttt{xarray} \citep{Hoyer_2017} to manage our data and model training. We also use \texttt{timm}~\citep{rw2019timm} for our ViT implementation.
All training is done on $10$ AMD EPYC 7313 CPU cores and one NVIDIA RTX A5000 GPU.
We leverage \texttt{fp16} floating point precision in our experiments. 

\subsection{Metrics} \label{sec:metrics}

We use the following definitions in our metric formulations
\begin{itemize}
    \item $N$ is the number of data points
    \item $H$ is the number of latitude coordinates.
    \item $W$ is the number of longitude coordinates.
    \item $X$ and $\Tilde{X}$ are the ground-truth and prediction, respectively.
\end{itemize}

The latitude weighting function is given by
\begin{equation}
    L(i) = \frac{\cos(H_i)}{\frac{1}{H}\sum_{i=1}^{H}\cos(H_i)}
\end{equation}

\subsubsection{Deterministic weather forecasting metrics}
\textbf{Root mean square error (RMSE) }
\begin{equation}
    \text{RMSE} = \frac{1}{N} \sum_{k=1}^{N} \sqrt{\frac{1}{H \times W} \sum_{i=1}^H \sum_{j=1}^W L(i)(\Tilde{X}_{k,i,j} - X_{k,i,j})^2}. \label{eq:lat_rmse}
\end{equation}

\textbf{Anomaly correlation coefficient (ACC) }
is the spatial correlation between prediction anomalies $\Tilde{X}^{'}$ relative to climatology and ground truth anomalies $X^{'}$ relative to climatology:
\begin{gather}
    \text{ACC} = \frac{\sum_{k,i,j} L(i) \Tilde{X}^{'}_{k,i,j} X^{'}_{k,i,j}}{\sqrt{\sum_{k,i,j} L(i) \Tilde{X}^{'2}_{k,i,j} \sum_{k,i,j} L(i) X^{'2}_{k,i,j}}}, \\
    \Tilde{X}^{'} = \Tilde{X}^{'} - C, X^{'} = X^{'} - C,
\end{gather}
in which climatology $C$ is the temporal mean of the ground truth data over the entire test set $C = \frac{1}{N}\sum_k X$.

\subsubsection{Probabilistic weather forecasting metrics}
\textbf{Spread-skill ratio (Spread by RMSE) } measures a probabilistic forecast's reliability. Let $N$ be the number of forecasts produced either by ensembling or drawing samples from a parametric prediction. Spread is given by
\begin{equation}
    \text{Spread} = \frac{1}{N}\sum_k^N\sqrt{\frac{1}{H \times W}\sum_{i=1}^{H}\sum_{j=1}^{W}L(i)\text{var}(\Tilde{X}_{i,j})}
\end{equation}

\textbf{Continuous ranked probability score } measures a probabilistic forecast's calibration and sharpness. Let $F$ denote the CDF of the forecast distribution. For a Gaussian distribution parameterized by mean $\mu$ and standard deviation $\sigma$, the closed-form, differentiable solution is
\begin{equation}
    \text{CRPS}(F_{\mu,\sigma},X)=\sigma\left\{\frac{X-\mu}{\sigma}\left[2\Phi\left(\frac{X-\mu}{\sigma}\right)-1\right]+2\phi\left(\frac{X-\mu}{\sigma}\right)-\frac{1}{\sqrt{\pi}}\right\}
\end{equation}
where $\Phi$ and $\phi$ are the CDF and PDF of the standard normal distribution, respectively.

\subsubsection{Climate downscaling metrics}
\textbf{Root mean square error (RMSE) }
This is the same as Equation~(\ref{eq:lat_rmse}). 

\textbf{Mean bias }
measures the difference between the spatial mean of the prediction and the spatial mean of the ground truth. A positive mean bias shows an overestimation, while a negative mean bias shows an underestimation of the mean value.
\begin{equation}
    \text{Mean bias} = \frac{1}{N \times H \times W} \sum_{k=1}^N \sum_{i=1}^H \sum_{j=1}^W \Tilde{X} - \frac{1}{N \times H \times W} \sum_{k=1}^N \sum_{i=1}^H \sum_{j=1}^W X
\end{equation}

\textbf{Pearson coefficient }
measures the correlation between the prediction and the ground truth. We first flatten the prediction and ground truth, and compute the metric as follows:
\begin{equation}
    \rho_{\Tilde{X}, X}=\frac{\operatorname{cov}(\Tilde{X}, X)}{\sigma_{\Tilde{X}} \sigma_X}
\end{equation}

\textbf{Masking for PRISM } Since PRISM does not record data over the oceans, we mask out those values for evaluation. Concretely, we set \texttt{NaN} values in the ground truth data to $0$. Then, we multiply the model's predictions by a binary mask that is $0$ wherever the ground truth data is originally \texttt{NaN} and is $1$ everywhere else.

\subsubsection{Climate projection metrics}
\textbf{Normalized spatial root mean square error (NRMSE$_s$) }
measures the spatial discrepancy between the temporal mean of the prediction and the temporal mean of the ground truth:
\begin{equation}
    \text{NRMSE}_s = \sqrt{\left\langle \left(\frac{1}{N} \sum_{k=1}^N \Tilde{X} - \frac{1}{N} \sum_{k=1}^N X \right)^2 \right\rangle} \bigg/ \frac{1}{N} \sum_{k=1}^N \left\langle X \right\rangle,
\end{equation}
in which $\langle A \rangle$ is the global mean of $A$:
\begin{equation}
    \langle A \rangle = \frac{1}{H \times W} \sum_{i=1}^H \sum_{j=1}^W L(i) A_{i,j}
\end{equation}

\textbf{Normalized global root mean square error (NRMSE$_g$) }
measures the discrepancy between the global mean of the prediction and the global mean of the ground truth:
\begin{equation}
    \text{NRMSE}_g = \sqrt{\frac{1}{N} \sum_{k=1}^N \left(\langle \Tilde{X} \rangle - \langle X \rangle\right)^2}  \bigg/ \frac{1}{N} \sum_{k=1}^N \left\langle X \right\rangle.
\end{equation}

\textbf{Total normalized root mean square error (Total) }
is the weighted sum of NRMSE$_s$ and NRMSE$_g$:
\begin{equation}
    \text{Total} = \text{NRMSE}_s + \alpha \cdot \text{NRMSE}_g,
\end{equation}
where $\alpha$ is chosen to be $5$ as suggested by~\citet{Watson-Parris2022}.

\section{Additional experiments}
\label{sec:additional_experiments}

\subsection{Climate projection} \label{sec:climate_proj}
\textbf{Task }
We consider the task of predicting the annual mean distributions of $4$ target variables in ClimateBench~\citep{Watson-Parris2022}: surface temperature, diurnal temperature range, precipitation, and the $90$th percentile of precipitation. 

\textbf{Baselines }
We compare ResNet, UNet, and ViT, three deep learning models supported by \name{} with CNN-LSTM, the deep learning baseline in ClimateBench. The network architectures of the three models are identical to Appendix~\ref{sec:network_arc}.

\textbf{Data }
We regrid the original ClimateBench data to $5.625^\circ$ for easy training and evaluation. The input variables include $4$ forcing factors: carbon dioxide (CO$_2$), sulfur dioxide (SO$_2$), black carbon (BC), and methane (CH$_4$). Similar to the deep learning baseline in ClimateBench, we stack $10$ consecutive years to predict the target variables of the current year. We standardize the input channels to have $0$ mean and $1$ standard deviation, but do not standardize the output variables. Training and validation data includes the historical data, ssp126, ssp370, ssp585, and the historical data with aerosol (hist-aer) and greenhouse gas (hist-GHG) forcings, and test data includes ssp245. We split train/validation data with a ratio of $0.9/0.1$.

\textbf{Training and evaluation } We train one network for each target variable. We use the same optimizer and scheduler as in Section~\ref{sec:forecast_benchmark}. We train for $50$ epochs with $16$ batch size, and use early stopping with a patience of $5$ epochs. We use mean-squared error as the loss function and evaluation metric. We report normalized spatial root mean square error (NRMSE$_s$), normalized global root mean square error (NRMSE$_g$), and Total = NRMSE$_s$ + 5$\times$NRMSE$_g$ as test metrics.

\textbf{Results }
Table~\ref{tab:climate_bench} shows the performance of different baselines on ClimateBench. CNN-LSTM and UNet are the best-performing methods, with each achieving the best performance in $5/12$ metrics, followed by ResNet which performs best on $2/12$ metrics. ViT achieves a reasonable performance but underperforms the CNN-based methods.

\begin{table}[h]
\centering
\caption{Performance of different deep learning baselines on ClimateBench. CNN-LSTM result is taken from ClimateBench.}
\label{tab:climate_bench}
\resizebox{1.0\textwidth}{!}{%
\begin{tabular}{@{}lcccccccccccc@{}}
\toprule
         & \multicolumn{3}{c}{Surface temperature}                                                   & \multicolumn{3}{c}{Diurnal temperature range}                                             & \multicolumn{3}{c}{Precipitation}                                                         & \multicolumn{3}{c}{$90$th percentile precipitation}                                       \\ \cmidrule(l){2-13} 
         & \multicolumn{1}{l}{NRMSE$_s$} & \multicolumn{1}{l}{NRMSE$_g$} & \multicolumn{1}{l}{Total} & \multicolumn{1}{l}{NRMSE$_s$} & \multicolumn{1}{l}{NRMSE$_g$} & \multicolumn{1}{l}{Total} & \multicolumn{1}{l}{NRMSE$_s$} & \multicolumn{1}{l}{NRMSE$_g$} & \multicolumn{1}{l}{Total} & \multicolumn{1}{l}{NRMSE$_s$} & \multicolumn{1}{l}{NRMSE$_g$} & \multicolumn{1}{l}{Total} \\ \midrule
CNN-LSTM & $0.107$                       & $0.044$                       & $\bold{0.327}$                   & $9.917$                       & $1.372$                       & $16.778$                  & $\bold{2.128}$                       & $0.209$                       & $\bold{3.175}$                   & $\bold{2.610}$                       & $0.346$                       & $\bold{4.339}$                   \\
ResNet   & $0.182$                       & $\bold{0.042}$                       & $0.395$                   & $9.128$                       & $\bold{0.737}$                       & $12.810$                  & $2.930$                       & $0.180$                       & $3.828$                   & $3.413$                       & $0.286$                       & $4.845$                   \\
UNet     & $\bold{0.097}$                       & $0.046$                       & $0.328$                   & $\bold{6.300}$                       & $0.946$                       & $\bold{11.030}$                  & $2.483$                       & $\bold{0.141}$                       & $3.187$                   & $3.122$                       & $\bold{0.282}$                       & $4.532$                   \\
ViT      & $0.191$                       & $0.092$                       & $0.650$                   & $7.725$                       & $0.746$                       & $11.460$                  & $2.909$                       & $0.327$                       & $4.545$                   & $3.615$                       & $0.418$                       & $5.704$                   \\ \bottomrule
\end{tabular}
}
\end{table}

\subsection{Extreme weather prediction} \label{sec:extreme_weather_appendix}
Table~\ref{tab:appendix_extreme_era5_forecasting} shows the performance of different models across various different lead times on the default test split and Extreme-ERA5. As discussed in Section~\ref{sec:extreme_events}, the performance of all models except Climatology is better on the extreme split than on the default split. 

\begin{table}
\vspace{-0.25in}
    \centering
    \caption{Latitude-weighted RMSE on the normal and extreme test splits of ERA5 for different lead times.}
    \label{tab:appendix_extreme_era5_forecasting}
    \resizebox{0.8\textwidth}{!}{\begin{tabular}{@{}lccccc@{}}
    \toprule
    T2M & 6 Hours     & 1 Day       & 3 Days & 5 Days & 10 Days\\ \midrule
    Climatology     & $5.87$ / $6.51$ & $5.87$ / $6.53$ & $5.87$ / $6.58$ & $5.88$ / $6.65$ & $5.89$ / $6.76$\\
    Persistence     & $2.76$ / $2.99$ & $2.13$ / $1.78$ & $2.99$ / $2.42$ & $3.26$ / $2.61$ & $3.59$ / $2.89$\\
    ResNet          & $0.72$ / $\bold{0.72}$ & $0.94$ / $\bold{0.91}$ & $1.50$ / $\bold{1.33}$ & $2.20$ / $\bold{1.86}$	& $2.78$ / $2.39$\\
    U-Net            & $0.76$ / $0.77$ & $1.04$ / $0.99$ & $1.65$ / $1.43$ & $2.26$ / $1.88$	& $2.76$ / $2.44$\\
    ViT            & $0.78$ / $0.80$  & $1.09$ / $1.05$ & $1.71$ / $1.55$ & $2.38$ / $2.04$	& $2.78$ / $\bold{2.30}$\\ \bottomrule
    \end{tabular}}
\end{table}

\begin{table}[t]
\centering
\caption{Performance of different models trained on one dataset (columns) and evaluated on another (rows). Training data for CMIP6 is available from the years $1850-2010$, at a 6 hour frequency. Training data for ERA5 is available from years $1979-2010$, at an one hour frequency.}
\label{tab:robustness_forecasting_entire}
\resizebox{0.8\textwidth}{!}{
\begin{tabular}{@{}cclcccccccc@{}}
\toprule
& & & \multicolumn{4}{c}{3 Days} & \multicolumn{4}{c}{5 Days} \\ \midrule
& & & \multicolumn{2}{c}{ERA5} & \multicolumn{2}{c}{CMIP6} & \multicolumn{2}{c}{ERA5} & \multicolumn{2}{c}{CMIP6} \\ \midrule
\multicolumn{1}{l}{} &  &  & \multicolumn{1}{l}{ACC} & \multicolumn{1}{l}{RMSE} & \multicolumn{1}{l}{ACC} & \multicolumn{1}{l}{RMSE} & \multicolumn{1}{l}{ACC} & \multicolumn{1}{l}{RMSE} & \multicolumn{1}{l}{ACC} & \multicolumn{1}{l}{RMSE}\\ \midrule
\multirow{9}{*}{ERA5} & \multirow{3}{*}{ResNet} & Z500 & $0.95$ & $315.07$ & $\bold{0.96}$ & $\bold{302.08}$ & $0.77$ & $646.57$ & $\bold{0.86}$ & $\bold{531.47}$\\
                   & & T850 & $\bold{0.93}$ & $\bold{1.84}$ & $0.91$ & $2.08$ & $0.80$ & $3.00$ & $\bold{0.83}$ & $\bold{2.77}$ \\
                   & & T2m  & $\bold{0.95}$ & $\bold{1.56}$ & $0.94$ & $1.85$ & $0.89$ & $2.35$ & $\bold{0.90}$ & $\bold{2.29}$\\ \cmidrule{2-11}
                   & \multirow{3}{*}{U-Net} & Z500 & $0.92$ & $388.17$ & $\bold{0.94}$ & $\bold{337.34}$ & $0.74$ & $686.90$ & $\bold{0.82}$ & $\bold{590.80}$\\
                   & & T850 & $\bold{0.91}$ & $\bold{2.09}$ & $0.90$ & $2.17$ & $0.78$ & $3.10$ & $\bold{0.81}$ & $\bold{2.93}$ \\
                   & & T2m & $\bold{0.95}$ & $\bold{1.72}$ & $0.93$ & $1.89$ & $0.89$ & $2.38$ & $\bold{0.89}$ & $\bold{2.37}$ \\ \cmidrule{2-11}
                   & \multirow{3}{*}{ViT} & Z500 & $0.93$ & $380.22$ & $\bold{0.93}$ & $\bold{373.57}$ & $0.68$ & $749.82$ & $\bold{0.82}$ & $\bold{592.36}$\\
                   & & T850 & $\bold{0.91}$ & $\bold{2.08}$ & $0.89$ & $2.31$ & $0.75$ & $3.27$ & $\bold{0.80}$ & $\bold{2.97}$ \\
                   & & T2m & $\bold{0.94}$ & $\bold{1.73}$ & $0.92$ & $2.10$ & $0.88$ & $2.54$ & $\bold{0.88}$ & $\bold{2.52}$ \\ \midrule
\multirow{9}{*}{CMIP6} & \multirow{3}{*}{ResNet} & Z500 & $0.95$ & $351.48$ & $\bold{0.98}$ & $\bold{240.37}$ & $0.77$ & $701.20$ & $\bold{0.89}$ & $\bold{496.04}$ \\
                   & & T850 & $0.92$ & $2.09$ & $\bold{0.96}$ & $\bold{1.43}$ & $0.79$ & $3.19$ & $\bold{0.89}$ & $\bold{2.33}$ \\
                   & & T2m  & $0.94$ & $1.88$ & $\bold{0.97}$ & $\bold{1.32}$ & $0.88$ & $2.54$ & $\bold{0.94}$ & $\bold{1.87}$ \\ \cmidrule{2-11}
                   & \multirow{3}{*}{U-Net} & Z500 & $0.92 $& $425.23$ & $\bold{0.96}$ & $\bold{300.19}$ & $0.75$ & $732.39$ & $\bold{0.85}$ & $\bold{575.38}$\\
                   & & T850 & $0.90$ & $2.30$ & $\bold{0.95}$ & $\bold{1.67}$ & $0.78$ & $3.29$ & $\bold{0.87}$ & $\bold{2.57}$ \\
                   & & T2m & $0.93$ & $2.00$ & $\bold{0.96}$ & $\bold{1.46}$ & $0.88$ & $2.57$ & $\bold{0.93}$ & $\bold{1.99}$ \\ \cmidrule{2-11}
                   & \multirow{3}{*}{ViT} & Z500 & $0.93$ & $413.76$ & $\bold{0.95}$ & $\bold{341.58}$ & $0.68$ & $820.65$ & $\bold{0.85}$ & $\bold{577.24}$\\
                   & & T850 & $0.90$ & $2.25$ & $\bold{0.94}$ & $\bold{1.83}$ & $0.75$ & $3.48$ & $\bold{0.86}$ & $\bold{2.60}$ \\
                   & & T2m & $0.92$ & $2.15$ & $\bold{0.95}$ & $\bold{1.59}$ & $0.85$ & $2.88$ & $\bold{0.92}$ & $\bold{2.03}$\\ \bottomrule
                   
\end{tabular}}
\end{table}
\begin{table}[h!]
\centering
\caption{Performance of different models trained on one dataset (columns) and evaluated on another (rows). The training years and data availability frequency is same for both the datasets.}
\label{tab:robustness_forecasting_complete}
\resizebox{0.8\textwidth}{!}{
\begin{tabular}{@{}cclcccccccc@{}}
\toprule
& & & \multicolumn{4}{c}{3 Days} & \multicolumn{4}{c}{5 Days} \\ \midrule
& & & \multicolumn{2}{c}{ERA5} & \multicolumn{2}{c}{CMIP6} & \multicolumn{2}{c}{ERA5} & \multicolumn{2}{c}{CMIP6} \\ \midrule
\multicolumn{1}{l}{} &  &  & \multicolumn{1}{l}{ACC} & \multicolumn{1}{l}{RMSE} & \multicolumn{1}{l}{ACC} & \multicolumn{1}{l}{RMSE} & \multicolumn{1}{l}{ACC} & \multicolumn{1}{l}{RMSE} & \multicolumn{1}{l}{ACC} & \multicolumn{1}{l}{RMSE}\\ \midrule
\multirow{9}{*}{ERA5} & \multirow{3}{*}{ResNet} & Z500 & $\bold{0.95}$ & $\bold{322.86}$ & $0.94$ & $345.00$ & $\bold{0.79}$ & $\bold{624.20}$ & $0.78$ & $646.48$\\
                   & & T850 & $\bold{0.93}$ & $\bold{1.90}$ & $0.90$ & $2.21$ & $\bold{0.81}$ & $\bold{2.91}$ & $0.79$ & $3.11$\\
                   & & T2m  & $\bold{0.95}$ & $\bold{1.62}$ & $0.93$ & $1.94$ & $\bold{0.90}$ & $\bold{2.33}$ & $0.88$ & $2.55$ \\ \cmidrule{2-11}
                   & \multirow{3}{*}{U-Net} & Z500 & $\bold{0.92}$ & $\bold{401.08}$ & $0.91$ & $422.77$ & $\bold{0.74}$ & $\bold{685.75}$ & $0.73$ & $712.62$\\
                   & & T850 & $\bold{0.90}$ & $\bold{2.17}$ & $0.88$ & $2.42$ & $\bold{0.78}$ & $\bold{3.10}$ & $0.76$ & $3.29$ \\
                   & & T2m & $\bold{0.94}$ & $\bold{1.81}$ & $0.91$ & $2.19$ & $\bold{0.89}$ & $\bold{2.44}$ & $0.86$ & $2.73$ \\ \cmidrule{2-11}
                   & \multirow{3}{*}{ViT} & Z500 & $\bold{0.91}$ & $\bold{426.70}$ & $0.90$ & $444.14$ & $\bold{0.72}$ & $\bold{698.08}$ & $0.72$ & $720.15$ \\
                   & & T850 & $\bold{0.89}$ & $\bold{2.27}$ & $0.87$ & $2.51$ & $\bold{0.78}$ & $\bold{3.12}$ & $0.76$ & $3.31$ \\
                   & & T2m & $\bold{0.94}$ & $\bold{1.88}$ & $0.91$ & $2.19$ & $\bold{0.89}$ & $\bold{2.43}$ & $0.87$ & $2.69$ \\ \midrule
\multirow{9}{*}{CMIP6} & \multirow{3}{*}{ResNet} & Z500 & $0.95$ & $357.66$ & $\bold{0.96}$ & $\bold{306.86}$ & $0.79$ & $682.76$ & $\bold{0.81}$ & $\bold{637.85}$  \\
                   & & T850 & $0.91$ & $2.11$ & $\bold{0.94}$ & $\bold{1.70}$ & $0.81$ & $3.09$ & $\bold{0.84}$ & $\bold{2.82}$ \\
                   & & T2m  & $0.93$ & $1.91$ & $\bold{0.96}$ & $\bold{1.53}$ & $0.88$ & $2.51$ & $\bold{0.91}$ & $\bold{2.24}$ \\ \cmidrule{2-11}
                   & \multirow{3}{*}{U-Net} & Z500 & $0.92$ & $440.82$ & $\bold{0.93}$ & $\bold{401.89}$ & $0.74$ & $739.94$ & $\bold{0.76}$ & $\bold{712.67}$ \\
                   & & T850 & $0.89$ & $2.37$ & $\bold{0.92}$ & $\bold{2.05}$ & $0.78$ & $3.27$ & $\bold{0.81}$ & $\bold{3.04}$ \\
                   & & T2m & $0.91$ & $2.23$ & $\bold{0.94}$ & $\bold{1.74}$ & $0.86$ & $2.73$ & $\bold{0.90}$ & $\bold{2.31}$ \\ \cmidrule{2-11}
                   & \multirow{3}{*}{ViT} & Z500 & $0.91$ & $460.14$ & $\bold{0.92}$ & $\bold{430.63}$ & $0.73$ & $753.18$ & $\bold{0.75}$ & $\bold{727.88}$ \\
                   & & T850 & $0.89$ & $2.40$ & $\bold{0.91}$ & $\bold{2.15}$ & $0.77$ & $3.29$ & $\bold{0.80}$ & $\bold{3.06}$ \\
                   & & T2m & $0.91$ & $2.15$ & $\bold{0.94}$ & $\bold{1.82}$ & $0.87$ & $2.68$ & $\bold{0.90}$ & $\bold{2.32}$ \\ \bottomrule
                   
\end{tabular}}
\end{table}

\subsection{Dataset robustness} \label{sec:dataset_robustness}

Table~\ref{tab:robustness_forecasting_entire} shows the comparison of the performance for different models when trained on ERA5 and evaluated on CMIP6 and vice versa at $3$ and $5$ days of lead time. For the CMIP6 evaluation purposes, the models trained on ERA5 were slightly worse than the moodels trained on CMIP6. Surprisingly, for evaluating on ERA5, models trained on CMIP6 were comparable, if not slightly better to the ones trained on ERA5. These results are in line with results of \cite{Nguyen2023}, thus highlighting the dataset usefulness of CMIP6 over ERA5. Note that the data's raw size is roughly similar for both the datasets as despite the ERA5's temporal training range being 1979-2010 in this setup, it's data availability frequency is 1 hour compared to 6 hour in CMIP6.

To find out whether this superiority of CMIP6 over ERA5 is just a result of differences in temporal range, we conducted the similar study but with same dataset temporal characteristics (i.e. setting training years as 1979-2010 and subsampling the data at 6 hours). This time the results just for ResNet at 3 day lead time is shown in Table~\ref{tab:robustness_forecasting} and for all models at different lead times, is shown in Table~\ref{tab:robustness_forecasting_complete}. These results show that the performance is slightly worse for both the cases now. Thus showing that the performance improvement of training over CMIP6 than ERA5 is likely just the bigger temporal range.

\section{Visualizations}
\label{sec:visualizations}
\name{} provides visualization functionality to help with an intuitive understanding of model performance. Below is an example figure generated by \name{} for visualizing the quality of a model's forecast. Each row represents a distinct time in the test set. The leftmost column shows weather conditions at the time the model is making a prediction from. The next column shows the ground truth conditions at the forecast horizon. The next column shows the model's predictions. The last column shows the model's bias, and its per-pixel forecast error.

\begin{figure}[h]
    \centering
    \includegraphics[width=\textwidth]{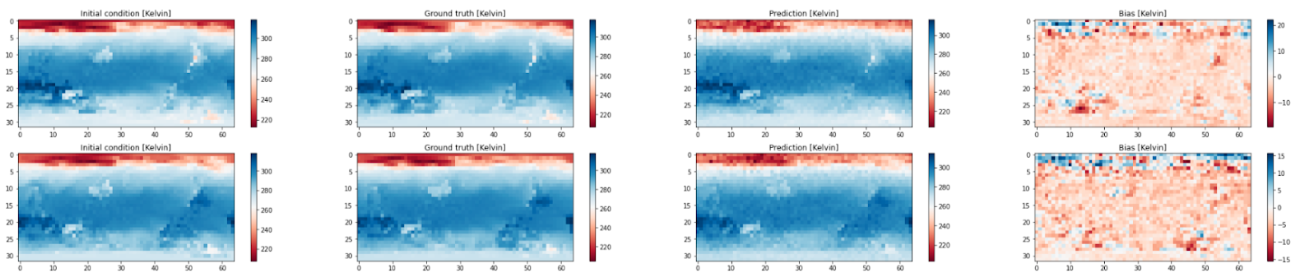}
    \caption{Example visualization of deterministic forecasting.}
\end{figure}

Additionally, \name{} can generate the rank histogram for probabilistic forecasts. A rank histogram that resembles a uniform distribution means that the ground truth value is indistinguishable from any member of the forecast ensemble. A rank histogram that is skew right occurs when the ground truth is consistently lower than the ensemble prediction. A rank histogram that appears U-shaped is indicative of both low biases and high biases. An example figure generated by \name{} for visualizing the rank histogram is shown below.

\begin{figure}[h]
    \centering
    \includegraphics[scale=0.5]{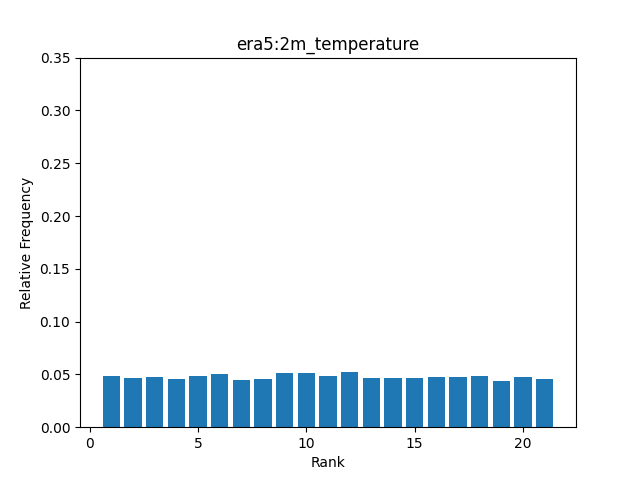}
    \caption{Example visualization of the rank histogram for probabilistic forecasting.}
\end{figure}

We show an example of how to generate another visualization called ``mean-bias'' in the next section.

\section{Code snippets} \label{sec:code_snippets}

\name{} can be used to download heterogeneous climate data from a variety of sources in a single function call. Here, we provide an example for downloading ERA5 2-meter temperature data at $5.625^\circ$ resolution from WeatherBench.

\begin{minted}
[
frame=lines,
framesep=2mm,
baselinestretch=1.2,
bgcolor=LightGray,
fontsize=\footnotesize,
linenos
]{python}
from climate_learn.data import download
download(
    root="./weatherbench-data",
    source="weatherbench",
    dataset="era5",
    resolution="5.625",
    variable="2m_temperature"
)
\end{minted}

Further, \name{} can process downloaded data into a form that is loadable into PyTorch. In fewer than \textbf{30 lines}, the following code loads raw ERA5 data; normalizes it; splits it into train, validation, testing sets; and prepares batches for the forecasting task.

\inputminted[
frame=lines,
framesep=2mm,
baselinestretch=1.2,
bgcolor=LightGray,
fontsize=\footnotesize,
linenos
]{python}{code/data-setup.py}

With the loaded data, \name{} can be used to build, train, and evaluate a model in fewer than \textbf{20 lines} of code.

\inputminted[
frame=lines,
framesep=2mm,
baselinestretch=1.2,
bgcolor=LightGray,
fontsize=\footnotesize,
linenos
]{python}{code/modeling.py}

\name{} can also be used to load pre-defined models (e.g., persistence, \citet{Rasp2021}) as follows.

\begin{minted}[
frame=lines,
framesep=2mm,
baselinestretch=1.2,
bgcolor=LightGray,
fontsize=\footnotesize,
linenos
]{python}
persistence = cl.load_forecasting_module(
    data_module=dm,
    preset="persistence"
)
rasp_theurey_2020 = cl.load_forecasting_module(
    data_module=dm,
    preset="rasp-theurey-2020"
)
\end{minted}

\name{} can use the trained forecasting models to produce visualizations in a single line of code. For example, one visualization of interest is the mean bias, which shows the expected error of the model's forecast, per pixel, over the evaluation period. 

\begin{minted}[
frame=lines,
framesep=2mm,
baselinestretch=1.2,
bgcolor=LightGray,
fontsize=\footnotesize,
linenos
]{python}
from climate_learn.utils.visualize import visualize_mean_bias
visualize_mean_bias(persistence, dm)
\end{minted}
\begin{figure}[h]
    \centering
    \includegraphics[scale=0.5]{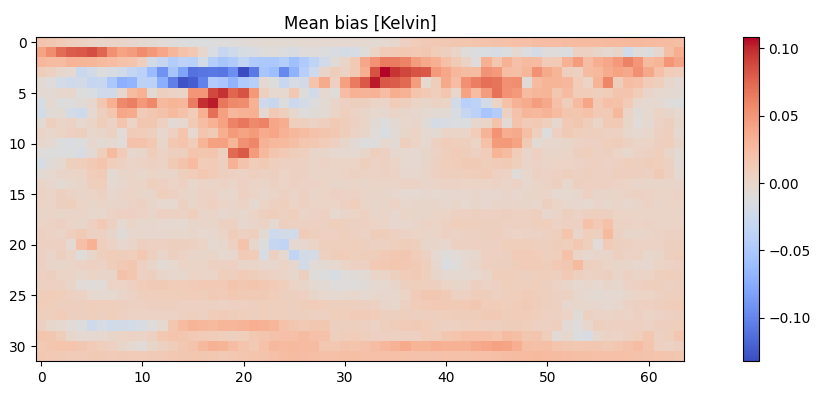}
    \caption{Visualization of the mean bias of temperature}
\end{figure}

This graphic shows that, on average, persistence has little bias below the equator. Over the northern part of North America, persistence achieves negative mean bias, which means it generally underpredicts 2-meter temperature in that region. Meanwhile, in the northern part of Europe, persistence achieves positives mean bias, indicating overprediction.